%% file: arxiv.tex
\documentclass{article}

\usepackage{cite}
\usepackage{amsmath,amssymb,amsfonts}
\usepackage{algorithmic}
\usepackage{graphicx}
\usepackage{textcomp}
\usepackage{xcolor}
\usepackage{xspace}

\usepackage[many]{tcolorbox}
\usepackage{hyperref}
\usepackage{cleveref}
\usepackage{bbding}
\usepackage{booktabs}
\usepackage{wrapfig}
\usepackage{float}

\def\BibTeX{{\rm B\kern-.05em{\sc i\kern-.025em b}\kern-.08em
    T\kern-.1667em\lower.7ex\hbox{E}\kern-.125emX}}

\def \VersionWithComments {}
\ifdefined \VersionWithComments
\usepackage{marginnote}
\newcommand{\marginX}{\marginnote{\huge{\quad\quad\textbf{!}\quad\quad}}}

\newcommand{\todo}[1]{\mbox{}{\color{blue}{\marginX{}\textbf{TODO}\ifx#1\\\else:\ \fi #1}}}
\fi
\usepackage{geometry}
\usepackage{authblk}

\usepackage{aics}
\usepackage{bxcoloremoji}
\usepackage{fontawesome5}

\title{CodeJudgeBench: Benchmarking LLM-as-a-Judge for Coding Tasks}
\author[1]{Hongchao Jiang}
\author[1]{Yiming Chen}
\author[1]{Yushi Cao}
\author[2]{Hung-yi Lee}
\author[1]{Robby T. Tan}

\affil[1]{ASUS Intelligent Cloud Services (AICS)}
\affil[2]{National Taiwan University}

\begin{document}

\thispagestyle{fancy}

\customabstract{
Large Language Models (LLMs) have significantly advanced the state-of-the-art in various coding tasks. Beyond directly answering user queries, LLMs can also serve as judges, assessing and comparing the quality of responses generated by other models. 
Such an evaluation capability is crucial both for benchmarking different LLMs and for improving response quality through response ranking.
However, despite the growing adoption of the LLM-as-a-Judge paradigm, its effectiveness in coding scenarios remains underexplored due to the absence of dedicated benchmarks.
To address this gap, we introduce CodeJudgeBench, a benchmark explicitly designed to evaluate the performance of LLM-as-a-Judge models across three critical coding tasks: code generation, code repair, and unit test generation.
Through comprehensive benchmarking of 26 LLM-as-a-Judge models, we find that recent thinking models significantly outperform non-thinking models on our carefully designed code judging tasks. Notably, even relatively small thinking models, such as Qwen3-8B, can outperform specially trained LLM-as-a-Judge models up to 70B in size. Nevertheless, all models still exhibit significant randomness in their judgment of coding tasks. For pairwise judging tasks, simply changing the order in which responses are presented can substantially impact accuracy. In addition, when judging code and unit tests written by different LLMs, LLM-as-a-Judge models also show variance in performance. This sensitivity raises concerns about the reliability and consistency of LLM-as-a-Judge in coding scenarios. Lastly, we study optimal prompting strategies for LLM-as-a-Judge. We find that using pair-wise comparison outperforms scalar point-wise judging. 
Furthermore, retaining comments and reasoning in the full, unprocessed LLM response leads to improved judge performance.

\vspace{0.5cm}
{\coloremojicode{1F917} \textbf{Dataset}: \href{https://huggingface.co/datasets/mattymchen/codejudgebench}{https://huggingface.co/datasets/mattymchen/codejudgebench}} \newline
{\faGithub \hspace{0.03cm} \textbf{GitHub}: \href{https://github.com/hongcha0/CodeJudgeBench}{https://github.com/hongcha0/CodeJudgeBench}}
}

\input{texs/introduction}
\input{texs/preliminary}

\input{texs/method}

\input{texs/experimentdesign}
\input{texs/results}
\input{texs/relatedwork}

\input{texs/conclusion}

\newpage

\bibliographystyle{plain}
\bibliography{ref}
\end{document}

%% file: texs/introduction.tex
\section{Introduction}

Large Language Models (LLMs)~\cite{gemini,Claude4,Claude37,gpt4,llama3} have significantly advanced the state-of-the-art in a wide range of automated software engineering tasks, including code generation~\cite{li2022competition}, code repair~\cite{self-debug}, and unit test generation~\cite{chen2023codet,mundler2024swt}. By harnessing their understanding of both natural language and programming constructs, LLMs have become indispensable tools for developers seeking automated coding assistance. As the volume of LLM-generated code continues to grow, there is an urgent need for scalable and reliable evaluation methods.
Recently, the LLM-as-a-Judge paradigm~\cite{gu2024survey,he2025code} has emerged as a promising solution for automating the assessment of code produced by both humans and machines. Unlike traditional automated metrics such as CodeBLEU~\cite{ren2020codebleu}, which depend on human-written reference implementations, LLM-as-a-Judge leverages the generative and evaluative capabilities of LLMs themselves to directly assess code quality, enabling more flexible and scalable evaluation pipelines.
% In summary, LLM-as-a-Judge offers several advantages. 
% First, it enables a scalable and robust evaluation of LLMs by establishing a general framework for comparing model outputs across diverse tasks and difficulty levels. Second, it facilitates automated response ranking, allowing systems to select the most accurate or optimal solution from a set of candidate responses, thereby improving overall performance.

% \input{tables/benchmark}
% \begin{figure}[t!] 
% \centering
% \includegraphics[width=\columnwidth]{imgs/model_performance.pdf}
% \caption{Comparison of LLM-as-a-Judge performance on previous benchmarks and proposed CodeJudgeBench (CodeGen Task).}
% \label{fig:method:previous}
% \end{figure}

% \begin{figure}[t!]
%     \centering

%     % Table on top (input from external file)
%     \begin{minipage}{\columnwidth}
%         \input{tables/benchmark}  % This should include \caption{} and \label{}
%     \end{minipage}

%     % \vspace{1em}

%     % Figure below
%     \begin{minipage}{\columnwidth}
%         \centering
%         \includegraphics[width=\columnwidth]{imgs/model_performance.pdf}
%         \caption{Overview of CodeJudgeBench and comparison of LLM-as-a-Judge performance on CodeGen Task with previous benchmarks.}
%         \label{fig:method:previous}
%     \end{minipage}
% \end{figure}

\begin{figure}[t]
    \centering
    % Table on the left
    \begin{minipage}[t]{0.48\textwidth}
        \vspace{-3cm}
        \input{tables/benchmark}
    \end{minipage}
    % \hfill
    % Image on the right
    \begin{minipage}[t]{0.48\textwidth}
        \centering
        \includegraphics[width=\textwidth]{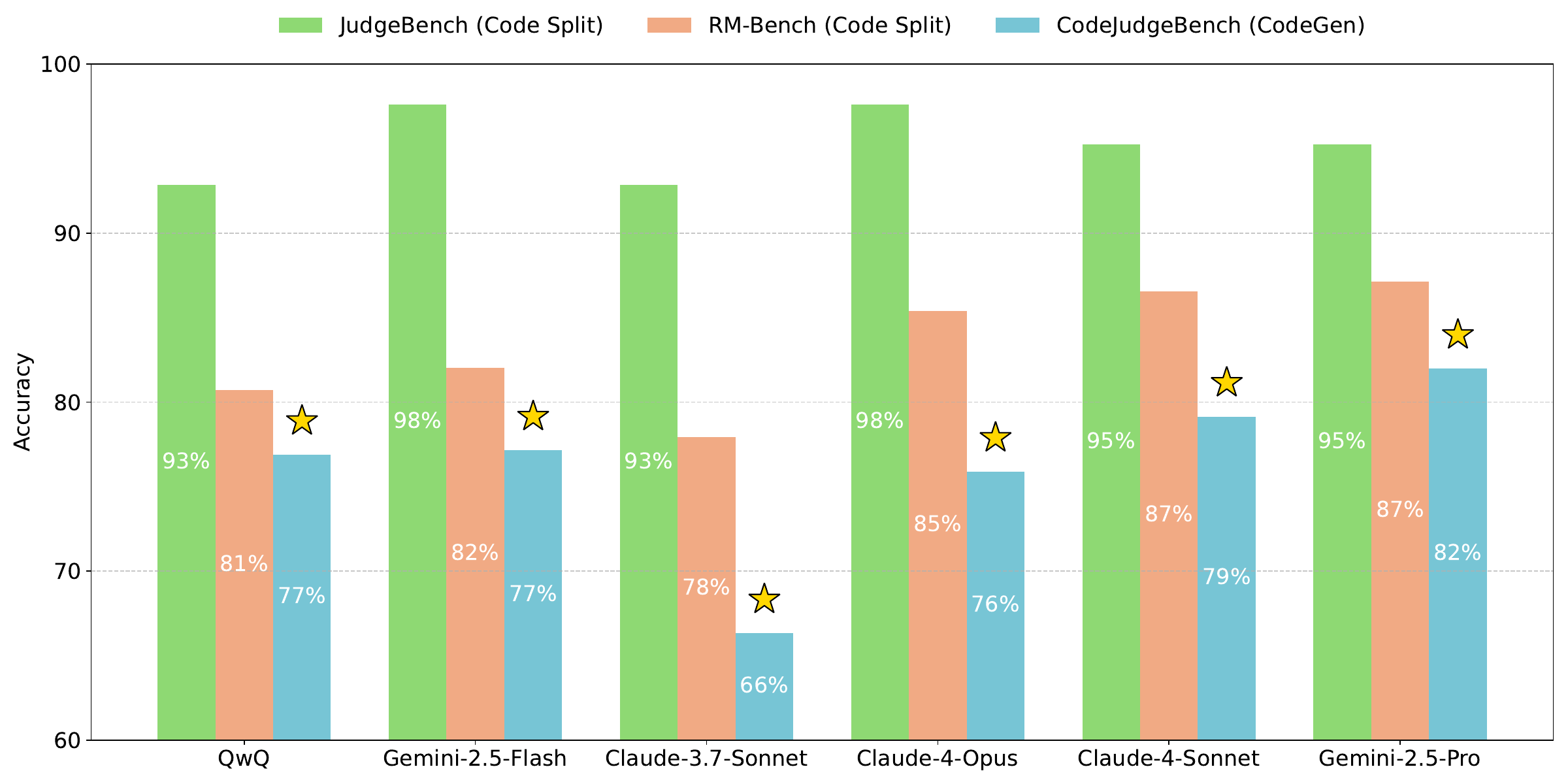}
    \end{minipage}
    \caption{Overview of CodeJudgeBench and comparison of LLM-as-a-Judge performance on CodeGen Task with previous benchmarks.}
    \label{fig:method:previous}
\end{figure}

Existing research can be broadly classified into two types of judging criteria: code functionality and code quality~\cite{he2025code}. Code quality assessment focuses on evaluating aspects such as readability and style, which closely align with human preferences~\cite{weyssow2024codeultrafeedback0,wang2025llms}.
In this work, we focus on execution-free judging of code functionality, specifically assessing whether the generated code or unit tests are functionally correct.
Execution-free judging determines correctness without code execution, avoiding the computational and operational challenges of managing execution environments and processing large numbers of solutions, which limit scalability during inference~\cite{le2018overfitting,zhou2024leveraging}.
Most existing LLM-as-a-Judge benchmarks~\cite{tan2025judgebench,liu2025rmbench} are designed for general domains and include only a small subset of relatively simple coding problems. 
Even benchmarks tailored for coding scenarios~\cite{zhao-etal-2025-codejudge,yang2025code0diting0,ficek2025scoring,zhou2025evaluating} tend to focus exclusively on code generation tasks, thereby overlooking the broader spectrum of coding activities that modern LLMs are increasingly capable of performing.

To address this gap, we introduce CodeJudgeBench, a benchmark consisting of 5,352 curated pairs. This represents a substantial increase in scale compared to previous benchmarks and encompasses tasks in code generation, code repair, and unit test generation.
We use state-of-the-art LLMs like Gemini-2.5-Pro and Claude-3.7-Sonnet to generate high-quality and challenging candidate responses.
The errors produced by these advanced LLMs are often subtle, resulting in fine-grained and nuanced differences between chosen and rejected responses. This makes the judgment process more challenging and requires a more thorough evaluation by the LLM-as-a-Judge.
In contrast, earlier benchmarks~\cite{tan2025judgebench,liu2025rmbench} often rely on weaker models like GPT-4o, which produce less challenging samples.
We compare the performance of various LLM-as-a-Judge models on code generation problems sampled from existing pairwise benchmarks and our proposed CodeJudgeBench (CodeGen Task).
As shown in Fig.~\ref{fig:method:previous}, all LLM-as-a-Judge models exhibit substantially lower accuracy on CodeJudgeBench compared to prior benchmarks. Frontier models such as Gemini-2.5-Pro achieve near-perfect accuracy on JudgeBench~\cite{tan2025judgebench}, indicating that these benchmarks are no longer adequate for tracking the rapid advancements of the latest models.

We benchmark a diverse set of LLMs on CodeJudgeBench, including both open-source and close-source models. 
In addition to general domain LLMs, we also evaluate models specifically tuned for coding or for LLM-as-a-Judge tasks~\cite{self-taught,pombal2025m}. 
Notably, we evaluate a new class of LLMs known as reasoning models, which are the current best-performing coding models. 
In this paper, we refer to reasoning models~\cite{deepseek-ai2025deepseekr1,qwen3,qwq32b,chen2025rm} that use long chain-of-thought~\cite{yeotong2025longcot} to enable capabilities like backtracking, self-verification, and reflection as \textit{thinking} models. 
Thinking models show increased performance gains with more tokens spent, allowing for effective inference-time scaling~\cite{snell2024scaling,ehrlich2025codemonkeys0}. 
Despite their recent popularity, it remains unclear how thinking models perform as LLM Judges, particularly for coding tasks. Importantly, strong code generation ability does not necessarily translate into strong code judgment capability~\cite{tan2025judgebench,zhao-etal-2025-codejudge}.

Overall, our main contributions include:
\begin{itemize}
    \item \textbf{Novel Benchmark:} We propose a challenging benchmark, CodeJudgeBench, tailored for evaluating LLM-as-a-Judge for code generation, code repair, and unit test generation.
    \item \textbf{Comprehensive Evaluation:} We evaluate the performance of 26 popular LLMs, revealing the capabilities of LLM-as-a-Judge on coding tasks more comprehensively.
    \item \textbf{Extensive Analysis:} By conducting various analysis experiments, we analyze the impact of different factors on LLM-as-a-Judge performance, providing valuable design suggestions for development.
\end{itemize}

%% file: tables/benchmark.tex
% \begin{table}[!t]
\centering
% \resizebox{\columnwidth}{!}{
% \begin{tabular}{lcccc}
% \toprule\toprule
%  & \textbf{CodeGen} & \multicolumn{1}{c}{\textbf{CodeRepair}} & \multicolumn{1}{c}{\textbf{TestGen}} & \textbf{\# Samples} \\
% \midrule
% JudgeBench{\tiny\cite{tan2025judgebench}} & \checkmark &   &   & 73 \\
% RM-Bench{\tiny\cite{liu2025rmbench}} & \checkmark &   &   & 684 \\
% CJ-Eval{\tiny\cite{zhao-etal-2025-codejudge}} & \checkmark &   &   & 1860 \\
% \midrule
% CodeJudgeBench & \checkmark & \checkmark & \checkmark  & 4260 \\
% \bottomrule\bottomrule
% \end{tabular}
% }

\resizebox{\columnwidth}{!}{
\begin{tabular}{llcccc}
\toprule\toprule
 & \textbf{Eval} & \textbf{CodeGen} & \multicolumn{1}{c}{\textbf{CodeRepair}} & \multicolumn{1}{c}{\textbf{TestGen}} & \textbf{\# Samples} \\
\midrule
JudgeBench{\tiny\cite{tan2025judgebench}} & Pairwise & \checkmark &   &   & 73 \\
RM-Bench{\tiny\cite{liu2025rmbench}} & Pairwise & \checkmark &   &   & 684 \\
CJ-Eval{\tiny\cite{zhao-etal-2025-codejudge}} & Pointwise & \checkmark &   &   & 1860 \\
CODE-DITING{\tiny\cite{yang2025code0diting0}} & Pointwise & \checkmark &   &   & 2952 \\
Scoring-Verifiers{\tiny\cite{ficek2025scoring}} & Ranking & \checkmark &   &   & 1680 \\
JETTS{\tiny\cite{zhou2025evaluating}} & BoN & \checkmark &   &   & 1682 \\
\midrule
CodeJudgeBench & Pairwise, BoN & \checkmark & \checkmark & \checkmark  & 5352 \\
\bottomrule\bottomrule
\end{tabular}
}

%% file: texs/preliminary.tex
\section{Preliminary}
 
In the LLM-as-a-Judge framework, an LLM is prompted to evaluate candidate responses based on their quality or correctness, eliminating the need for human validation. 

Formally, LLM-as-a-Judge is defined as:
$$
J \leftarrow LLM(p \oplus r \oplus  q),
$$
where $J$ is the final judgment or verdict produced by the LLM-as-a-Judge, $p$ is the programming task, $r$ is the response or set of responses to be evaluated, and $q$ is the instruction prompting the LLM to act as a judge. The operator $\oplus$ specifies the method for concatenating or formatting $p$, $r$, and $q$ into a single prompt for the LLM; the exact construction may differ across different LLM-as-a-Judge variants.

\begin{wrapfigure}{r}{0.5\textwidth}
\vspace{-0.3cm}
\centering
\includegraphics[width=\linewidth]{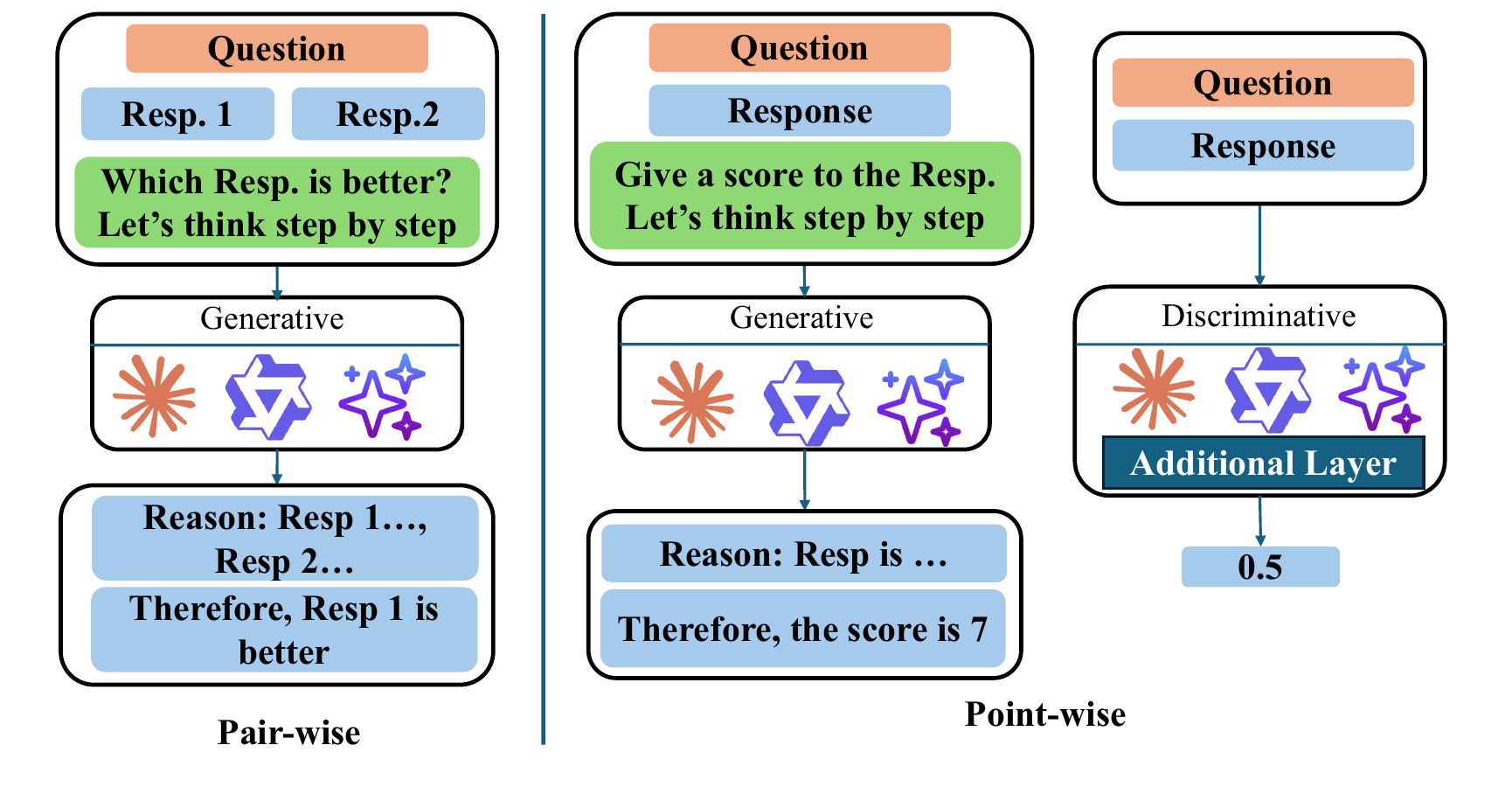} 
\caption{We benchmark three variants of LLM-as-a-Judge in our study.}
\label{fig:preli:judge_variant} 
\end{wrapfigure}

In this study, we examine three variants of LLM-as-a-Judge (as shown in Fig.~\ref{fig:preli:judge_variant}), which stem from two main approaches: pair-wise and point-wise~\cite{zheng2023judging}.

\paragraph{\textbf{Pair-wise LLM-as-a-Judge:}}
In the pair-wise LLM-as-a-Judge setting, the model receives a programming question prompt $p$, two candidate responses $r_1$ and $r_2$, and a query $q$ that asks the model to determine which response is preferable. In our study, we employ both thinking and Chain-of-Thought (CoT) LLM-as-a-Judge. These models first generate a rationale that analyzes both responses in the context of the prompt $p$, and then produce a final decision (e.g., response 1 is better).

\paragraph{\textbf{Point-wise LLM-as-a-Judge:}}
In the point-wise LLM-as-a-Judge setting, the model is provided with a programming question prompt $p$ and a single candidate response $r$. The associated query $q$ instructs the LLM-as-a-Judge to evaluate this response by assigning a score. For thinking and CoT LLMs, the model generates a rationale that analyzes the response before producing a final score, typically on a Likert scale (e.g., $one$ to $five$). In contrast to the generative approach, discriminative models add an additional layer atop the base model, which is fine-tuned via supervised learning to directly output a scalar score representing the quality of the response. Finally, the response with the highest score will be chosen as the final response.

%% file: texs/method.tex
\section{CodeJudgeBench}

\begin{figure*}[!htbp] 
\centering
\includegraphics[width=\linewidth]{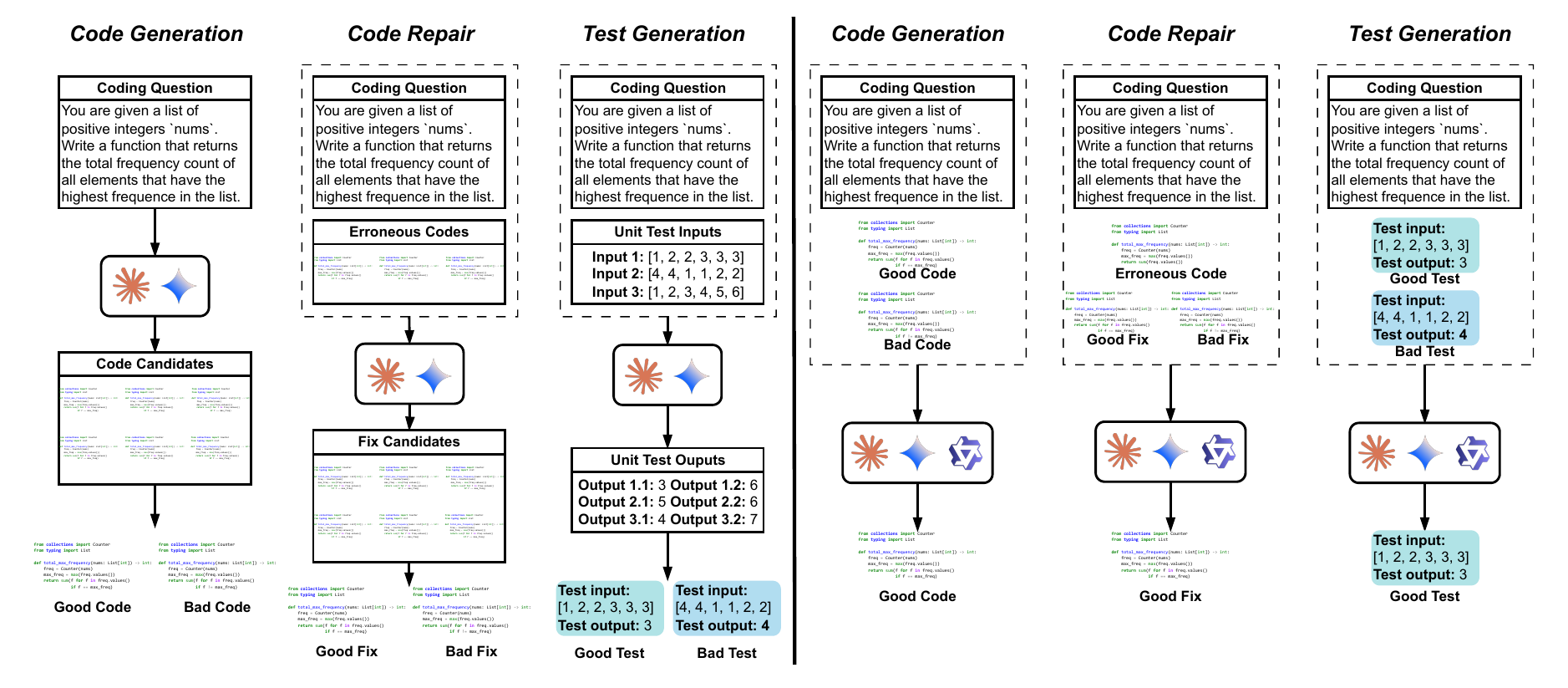} 
\caption{Overview of the proposed CodeJudgeBench. The left side illustrates the data curation process of CodeJudgeBench. The right side illustrates the evaluation process of CodeJudgeBench.}
\label{fig:method:tasks} 
\end{figure*}

\subsection{Overview}
As shown in Fig.~\ref{fig:method:tasks}, CodeJudgeBench evaluates the capabilities of LLM-as-a-Judge across three crucial coding tasks: code generation~\cite{10.1145/3672456}, code repair~\cite{tang2024code}, and test generation~\cite{10.1145/3663529.3663839}. 
Each data point in CodeJudgeBench consists of a triplet: {Instruction, Good Response, Bad Response}. 
The LLM-as-a-Judge is then tasked with evaluating both responses and choosing the one that better satisfies the instruction.
The construction of CodeJudgeBench involves three primary stages: response collection, response verification, and response pairing. 
We source challenging coding questions from LiveCodeBench~\cite{jain2025livecodebench}, which mitigates data contamination by continually collecting new problems from platforms such as LeetCode, AtCoder, and CodeForces.
We use LiveCodeBench-v6 comprising 1,055 challenging coding competition problems published between May 2023 and April 2025.
The subsequent sections provide further details of the construction process for each specific task.

\subsection{Code Generation (CodeGen)}

\noindent \textbf{Task Definition:}
In the Code Generation task, the LLM-as-a-Judge is provided with a coding problem statement and two candidate code snippets, and must determine which snippet is correct.

\noindent \textbf{Response Collection:}
We adopt the standard code generation setup, providing a detailed problem description along with illustrative input-output test cases. Multiple candidate code solutions are generated for each problem.

\noindent \textbf{Response Verification:}
To verify the correctness of the generated responses, we utilize the comprehensive suite of unit tests provided by LiveCodeBench. Responses that pass all unit tests are labeled as \textit{good}, while those that fail any test are labeled as \textit{bad}.

\noindent \textbf{Response Pairing:}
For each coding problem, we randomly select one good and one bad response to form an evaluation pair. 
In cases where repeated sampling yields all correct or all incorrect responses, we discard the corresponding coding problem from the evaluation set. 

\subsection{Code Repair (CodeRepair)}
\noindent \textbf{Task Definition:}
In the code repair task, the LLM-as-a-Judge is presented with the original coding problem statement, an erroneous code snippet along with its corresponding error message, and two candidate code repairs. 
The task of the LLM-as-a-Judge is to identify which candidate represents the correct fix.

\noindent \textbf{Response Collection:}
The erroneous code snippets are sourced from the incorrect responses identified in the code generation task.
Each incorrect snippet, together with its associated error message obtained from failed unit tests, is fed back into the coding models to generate repaired code.
Multiple repair candidates are produced for each erroneous code snippet.

\noindent \textbf{Response Verification:}
Each repair candidate is verified using unit tests; those passing all tests are labeled as good, while those failing any test are labeled as bad.

\noindent \textbf{Response Pairing:}
For each erroneous snippet, one good and one bad repair candidate are randomly paired to form an evaluation instance.
Problems for which only correct or only incorrect repairs are available are excluded from the evaluation set.

\subsection{Test Generation (TestGen)}

\noindent \textbf{Task Definition:}
Following prior work~\cite{chen2023codet}, this task focuses on generating unit tests directly from problem statements, without relying on any reference code.
The LLM-as-a-Judge is presented with a problem statement and two candidate unit test cases (input-output pairs), and must identify the correct test case.

\noindent \textbf{Response Collection:}
To construct candidate unit tests, the problem statement and a test input from the original dataset are provided, and the coding model is tasked with generating the expected output.

\noindent \textbf{Response Verification:}
Candidate outputs are verified by direct comparison with the ground-truth outputs from the dataset.

\noindent \textbf{Response Pairing:}
Each validated output (correct or incorrect) is paired with its input to form good and bad test cases. 
Problems for which only correct or only incorrect test responses are available are excluded from the evaluation set.
Finally, a correct and an incorrect unit test case, each with different test inputs, are randomly paired to form an evaluation instance.

\subsection{Data Statistics}
To ensure the high quality and diversity of the generated responses, we utilize three state-of-the-art LLMs—Claude-3.7-Sonnet, Gemini-2.5-Flash, and Gemini-2.5-Pro, all of which demonstrate strong performance on coding benchmarks. For the CodeGen task, we further include models such as Qwen3-235B, Claude-4-Sonnet, Claude-4-Opus, Gemini-2.5-Flash-Lite. Tab.~\ref{tab:data-statistics} summarizes the data statistics of CodeJudgeBench.

Following~\cite{zhang2025codecriticbench,frick2025how}, we categorize the samples in each task into three difficulty levels: easy, medium, and hard based on the proportion of LLMs that correctly judge each sample. As pairwise judging is a binary task susceptible to random guessing, we only use top performing LLMs from both open-source and close-source for the assessment.

\input{tables/statistics}

%% file: tables/statistics.tex
\begin{table*}[h!]
\centering
\resizebox{\textwidth}{!}{
\begin{tabular}{lccccccccccccc}
\toprule \toprule
\textbf{Source}  & \multicolumn{4}{c}{\textbf{Code Generation}}                       & \multicolumn{4}{c}{\textbf{Code Repair}}                           & \multicolumn{4}{c}{\textbf{Unit Test Generation}}                  & \textbf{Overall}     \\
\cmidrule(lr){2-5} \cmidrule(lr){6-9} \cmidrule(lr){10-13}
                 & \textbf{Easy} & \textbf{Medium} & \textbf{Hard} & \textbf{Overall} & \textbf{Easy} & \textbf{Medium} & \textbf{Hard} & \textbf{Overall} & \textbf{Easy} & \textbf{Medium} & \textbf{Hard} & \textbf{Overall} & \multicolumn{1}{l}{} \\
\midrule
Claude-3.7       & 162            & 92             & 71           & 325              & 385            & 262             & 231           & 878              & 66            & 69             & 171           & 306              & 1509                 \\
Gemini-2.5-Flash & 115            & 123             & 192           & 430              & 149             & 197             & 308           & 654              & 88            & 64             & 167           & 319              & 1403                 \\
Gemini-2.5-Pro   & 64             & 57              & 135           & 256              & 204            & 244             & 429           & 877              & 30            & 29              & 156           & 215              & 1348                 \\
Gemini-2.5-Flash-Lite & 119       & 114             & 156           & 389              & -              & -               & -             & -                & -             & -               & -             & -                & 389  \\
Qwen3-235B & 46 & 59 & 113 & 218 & - & - & - & - & - & - & - & - & 218  \\
Claude-4-Sonnet & 115 & 73 & 97 & 285 & - & - & - & - & - & - & - & - & 285  \\
Claude-4-Opus & 73 & 62 & 65 & 200 & - & - & - & - & - & - & - & - & 200  \\

\midrule
Overall          & 694            & 580             & 829           & 2103             & 738            & 703             & 968          & 2409             & 184           & 162             & 494           & 840              & 5352                \\
\bottomrule \bottomrule
\end{tabular}
}
\caption{Data statistics of CodeJudgeBench.}
\label{tab:data-statistics}
\end{table*}

%% file: texs/experimentdesign.tex
\section{Experiment Design}

\subsection{Research Questions}
Our work aims to answer the following three Research Questions (RQs).
\begin{itemize}
    \item \textbf{RQ1: How well does LLM-as-a-Judge perform on coding tasks?} In RQ1, we investigate the performance of a variety of LLM-as-a-Judge models on coding tasks (i.e., code generation, code repair, and unit test generation).
    \item \textbf{RQ2: How robust and generalizable is LLM-as-a-Judge?} In RQ2, we study whether current LLM-as-a-Judge models can generalize across different model responses and are robust against candidate position swap.
    \item \textbf{RQ3: How does prompting impact LLM-as-a-Judge performance?} In RQ3, we study the effect of different prompting formats, specifically point-wise and pair-wise evaluation, on the performance of LLM-as-a-Judge models. We also examine the impact of candidate response pre-processing by comparing three approaches: using the full response, retaining only code and comments, and using code only. Lastly, we explore the use of pair-wise prompting for inference-time scaling.
\end{itemize}

\subsection{Selected Baselines}

To evaluate the capabilities of LLM-as-a-Judge, we choose multiple representative LLMs, as shown in Tab.~\ref{tab:expdesign:llms}. We classify these models based on whether they are capable of thinking, open-source, and trained on domain-specific (i.e., code datasets) or task-specific data (i.e., LLM-as-a-Judge datasets). 
The details of these models are as follows:

\input{tables/models}

\begin{itemize}
    \item \textbf{Gemini}: Gemini-2.5~\cite{gemini}, an advanced iteration in the Gemini series of LLMs, including two specialized thinking variants: Gemini-2.5-Pro, optimized for coding tasks and complex questions, and Gemini-2.5-Flash, designed for rapid execution of complex tasks. Both are used in our experiments. To assess the importance of reasoning capabilities, we also include the non-thinking models Gemini-2.0-Flash and Gemini-2.0-Flash-Lite.
    \item \textbf{Claude}: Claude 3.7~\cite{Claude37} and Claude 4~\cite{Claude4} represent the latest advancements in Anthropic's Claude Sonnet family, offering powerful proprietary models designed for reasoning tasks, especially in coding tasks.
    In this work, we evaluate Claude-3.7-Sonnet, Claude-4-Sonnet, and Claude-4-Opus. For comparison, we also include the non-thinking model Claude-3.5-Sonnet-v2.

    \item \textbf{AceCodeRM}: AceCodeRM~\cite{AceCoder} is a point-wise discriminative LLM judge specifically trained for evaluating code. It is trained on 89K good and bad code pairs generated by GPT-4o. We evaluate both AceCodeRM-7B and AceCodeRM-32B.
    \item \textbf{Qwen2.5-Coder}: Qwen2.5-Coder~\cite{qwen2.5} is the code-tuned version of the Qwen2.5 LLM series, trained on 5.5 trillion tokens, including source code and synthetic data. We evaluate Qwen2.5-Coder-32B-Instruct, the best performing model in the series, which excels in code generation, code reasoning, and code fixing.
    \item \textbf{Skywork-Critic}: Skywork-Critic~\cite{skyworkcritic2024} is a series of LLM Judges developed by the SkyworkAI team that excel at pair-wise evaluation. We evaluate the largest and best-performing model, Skywork-Critic-70B, which is fine-tuned from Llama3.1-70B~\cite{llama3}.
    \item \textbf{Prometheus}: Prometheus~\cite{kim2024prometheus,pombal2025m} is a suite of open-source LLM judges that can provide both point-wise and pair-wise judgments based on user-defined score rubrics. We evaluate the latest and best-performing iteration of Prometheus, Prometheus-14B~\cite{pombal2025m}.
    \item \textbf{Self-Taught}: The Self-Taught evaluator~\cite{self-taught}, developed by Meta, is an LLM-as-a-Judge trained iteratively using synthetic data without human annotations. Llama3.1-70B~\cite{llama3} undergoes self-training on self-generated reasoning traces and final judgments, continuously improving its LLM-as-a-Judge capabilities with each iteration.

    \item \textbf{R1-Distill}: R1-Distill~\cite{deepseek-ai2025deepseekr1} is a series of models released by DeepSeek, which are distilled from DeepSeek R1. We evaluate DeepSeek-R1-Distill-Qwen-14B/32B, which are distilled using Qwen2.5-14B and Qwen2.5-32B. DeepSeek-R1-0528-Qwen3-8B uses Qwen3-8B and is distilled from the latest version of DeepSeek R1, DeepSeek-R1-0528.
    \item \textbf{Qwen3}: Qwen3~\cite{qwen3}, the latest installment in the Qwen LLM series, features a range of open-source models with different parameter sizes, delivering state-of-the-art performance across multiple tasks and domains. We evaluate the Qwen3-8B, 14B, and 32B models. 
    \item \textbf{QwQ}: QwQ-32B~\cite{qwq32b} is the reasoning model of the Qwen series. It is specifically trained for deep thinking and complex reasoning, capable of achieving competitive performance against state-of-the-art reasoning models like DeepSeek-R1 and o1-mini.
    \item \textbf{RM-R1}: RM(Reward Model)-R1~\cite{chen2025rm} is a pair-wise reasoning LLM-as-a-Judge model that uses a chain-of-rubrics mechanism. Rubrics are dynamically generated at the sample-level based on the specific domain (e.g., chat or math/code), and candidate responses are evaluated against these self-generated rubrics. We evaluate both RM-R1-14B and RM-R1-32B which are trained from DeepSeek-R1-Distilled-Qwen-14B and DeepSeek-R1-Distilled-Qwen-32B respectively.
    \item \textbf{DeepCoder}: DeepCoder-14B-Preview~\cite{deepcoder2025} is a specialized reasoning LLM fine-tuned from DeepSeek-R1-Distilled-Qwen-14B, with a focus on code generation. Despite its 14B parameter size, it delivers performance comparable to OpenAI's o3-mini on LiveCodeBench.
    \item \textbf{Phi-4}: Phi-4~\cite{abdin2024phi}, developed by Microsoft, is a series of small reasoning models trained on high-quality synthetic and public data. We evaluate Phi4-Reasoning-Plus, a 14B model fine-tuned from Phi-4 using supervised fine-tuning on chain-of-thought traces and reinforcement learning, with an emphasis on math, science, and coding skills.
    \item \textbf{AceReason-Nemotron}: AceReason-Nemotron~\cite{chen2025acereason0nemotron0}, developed by Nvidia, is a math and code reasoning model trained using RL. 
    We evaluate AceReason-Nemotron-14B, which is trained from DeepSeek-R1-Distilled-Qwen-14B. It is first trained on math-only prompts, then on code-only prompts.

\end{itemize}

\subsection{Impelmentation Details}
For the Judge-Tuned LLMs, we follow the prompts and sampling parameters provided in their official implementations. In the case of general LLMs, we use the pair-wise prompt from~\cite{tan2025judgebench}, which instructs the LLM to first generate its own reference answer, which is then used to compare and evaluate candidate responses. The LLM is instructed to choose the better response without allowing for ties. 
We examine the impact of postprocessing and the differences between point-wise and pair-wise evaluation in~\Cref{sec:rq3}. To mitigate the risk of random guessing, each sample pair is evaluated twice. The good response is alternately placed as the first (i.e., position A) and second (i.e., position B) candidate, and the results are averaged.
The effect of candidate ordering is further examined in~\Cref{sec:rq2}.

% \subsection{Data Curation Setup}
% We use Gemini-2.5-Pro, Gemini-2.5-Flash, and Claude Sonnet 3.7 as LLM programmers to generate candidate responses. All three LLMs are thinking models, capable of handling complex coding tasks. 
% During sampling, we set the temperature to 1.0 to encourage response diversity. 
% To ensure output quality, we allow each model to generate responses up to its maximum output length. 
% For each coding problem, we sample ten responses from each model and use unit tests to categorize them into good and bad candidate pools. The pair instance is then created by randomly selecting one response from each pool.

%% file: tables/models.tex
\begin{table}[!htbp]
\centering
% \resizebox{\columnwidth}{!}{
\begin{tabular}{lcccc}
\toprule\toprule
\textbf{Model} & \textbf{Thinking} & \textbf{Open-Source} & \textbf{Code-Tuned} & \textbf{Judge-Tuned} \\
\midrule
Claude-3.5 &  & & & \\
Claude-3.7 & $\checkmark$ & & & \\
Claude-4 & $\checkmark$ & & & \\
Gemini-2.0 & & & & \\
Gemini-2.5 & $\checkmark$ & & & \\
R1-Distill & $\checkmark$ & $\checkmark$ & & \\
RM-R1 & $\checkmark$ & $\checkmark$ & & $\checkmark$ \\
QwQ & $\checkmark$ & $\checkmark$ & & \\
Qwen3 & $\checkmark$ & $\checkmark$ & & \\
AceReason-Nemotron & $\checkmark$ & $\checkmark$ & $\checkmark$ & \\
DeepCoder & $\checkmark$ & $\checkmark$ & $\checkmark$ & \\
Qwen2.5-Coder & & $\checkmark$ & $\checkmark$ & \\
Phi-4 & $\checkmark$ & $\checkmark$ & & \\
AceCodeRM  & & $\checkmark$ & $\checkmark$ & $\checkmark$ \\
Self-Taught  &  & $\checkmark$ & & $\checkmark$ \\
Skywork-Critic  &  & $\checkmark$ & & $\checkmark$ \\
Prometheus  &  & $\checkmark$ & & $\checkmark$ \\

\bottomrule \bottomrule
\end{tabular}
% }
\caption{Summary of LLM-as-a-Judge evaluated on CodeJudgeBench}
\label{tab:expdesign:llms}
\end{table}

%% file: texs/results.tex
\clearpage
\section{Results}

\subsection{RQ1: How well does LLM-as-a-Judge perform on coding tasks?}
Tab.~\ref{tab:result:main} presents the performance of various LLM-as-a-Judge models on CodeJudgeBench tasks. Overall, thinking models—such as DeepCoder-14B, AceReason-14B, Qwen3, QwQ, RM-R1, Claude 3.7/4, and Gemini-2.5—Pro/Flash consistently outperform others. These models allocate more tokens for code analysis, which enhances their ability to understand and accurately judge code responses. Notably, smaller thinking models, such as Qwen3-8B, surpass CoT models like Prometheus-14B and Self-Taught 70B in overall accuracy.
In contrast, non-thinking models—including proprietary models like Claude-3.5 and models specifically fine-tuned for LLM-as-a-Judge tasks such as Prometheus-14B—achieve accuracies below 60\%, approaching the random guess baseline of 50\%. The superior performance of thinking models is largely attributable to their self-verification capabilities, which are crucial for effective LLM-as-a-Judge systems.

\input{tables/main_results}

Interestingly, fine-tuning thinking models specifically for LLM-as-a-Judge tasks does not always yield improved performance. For example, RM-R1 underperforms relative to similarly sized models such as Qwen3-32B and QwQ, likely due to insufficient code-related training data in LLM-as-a-Judge datasets, which often focus on modeling general human preferences. Among all evaluated models, closed-source models such as Gemini-2.5-Pro and Claude-4-Sonnet achieve the highest scores on CodeJudgeBench.

In terms of task difficulty, judging the correctness of unit test generation is the most challenging for LLM-as-a-Judge models, followed by code generation, with code repair being the easiest. This may be because code generation and code repair are more common tasks and thus more frequently encountered during training, whereas test generation is less prevalent. Furthermore, code generation and code repair provide LLM-as-a-Judge models with richer contextual information, such as code snippets and error messages, which facilitates more accurate judgment. In contrast, test generation provides only the problem statement, making evaluation inherently more difficult.
While larger model sizes often correlate with improved performance, this trend is not as apparent in CodeJudgeBench. Several 14B models perform comparably to their larger counterparts; for example, RM-R1 14B achieves results similar to RM-R1 32B, and Qwen3-14B is on par with Qwen3-32B.

\begin{tcolorbox}[colback=paleviolet, size=title]
\textbf{\textcolor{titlecolor}{Answer to RQ1}}: 
We observe that earlier non-thinking and CoT LLM-as-a-Judge models struggle to accurately identify the correct response in coding tasks. In contrast, the latest thinking LLM-as-a-Judge models demonstrate significantly higher performance. Interestingly, recent efforts to fine-tune thinking LLMs specifically for LLM-as-a-Judge tasks do not yield improvements over general-purpose thinking models. These findings suggest that future research should focus on developing more effective approaches for training and selecting coding-specific judges.
\end{tcolorbox}

\vspace{0.6cm}
\subsection{RQ2: How robust and generalizable is LLM-as-a-Judge?}
\label{sec:rq2}
Ideally, LLM-as-a-Judge models should be capable of evaluating a wide range of outputs, with their assessment remaining unaffected by superficial factors such as response ordering or model-specific characteristics. Motivated by this, we study the robustness and generalization abilities of LLM judges in two key settings: (1) the impact of response ordering in pair-wise evaluation, and (2) the variation in performance when judging responses generated by different models.

\begin{figure*}[!htbp]
    \centering
    \includegraphics[width=0.9\linewidth]{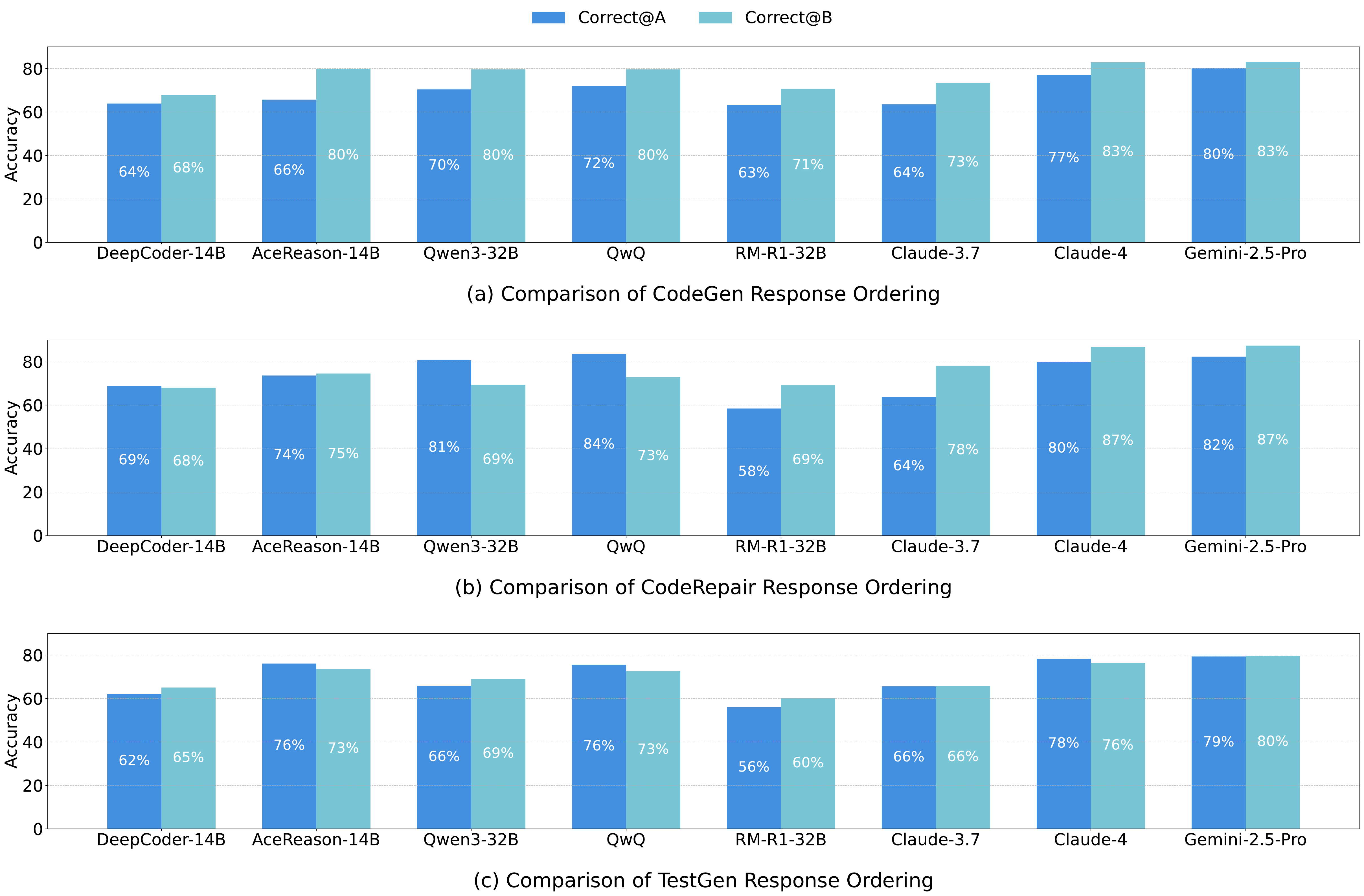}
    \caption{The performance of LLM-as-a-Judge when the correct response is presented in either position A or position B.}
    \label{fig:gen-study}

    \vspace{1em} % Optional: adds vertical space between the figures

    \includegraphics[width=0.9\linewidth]{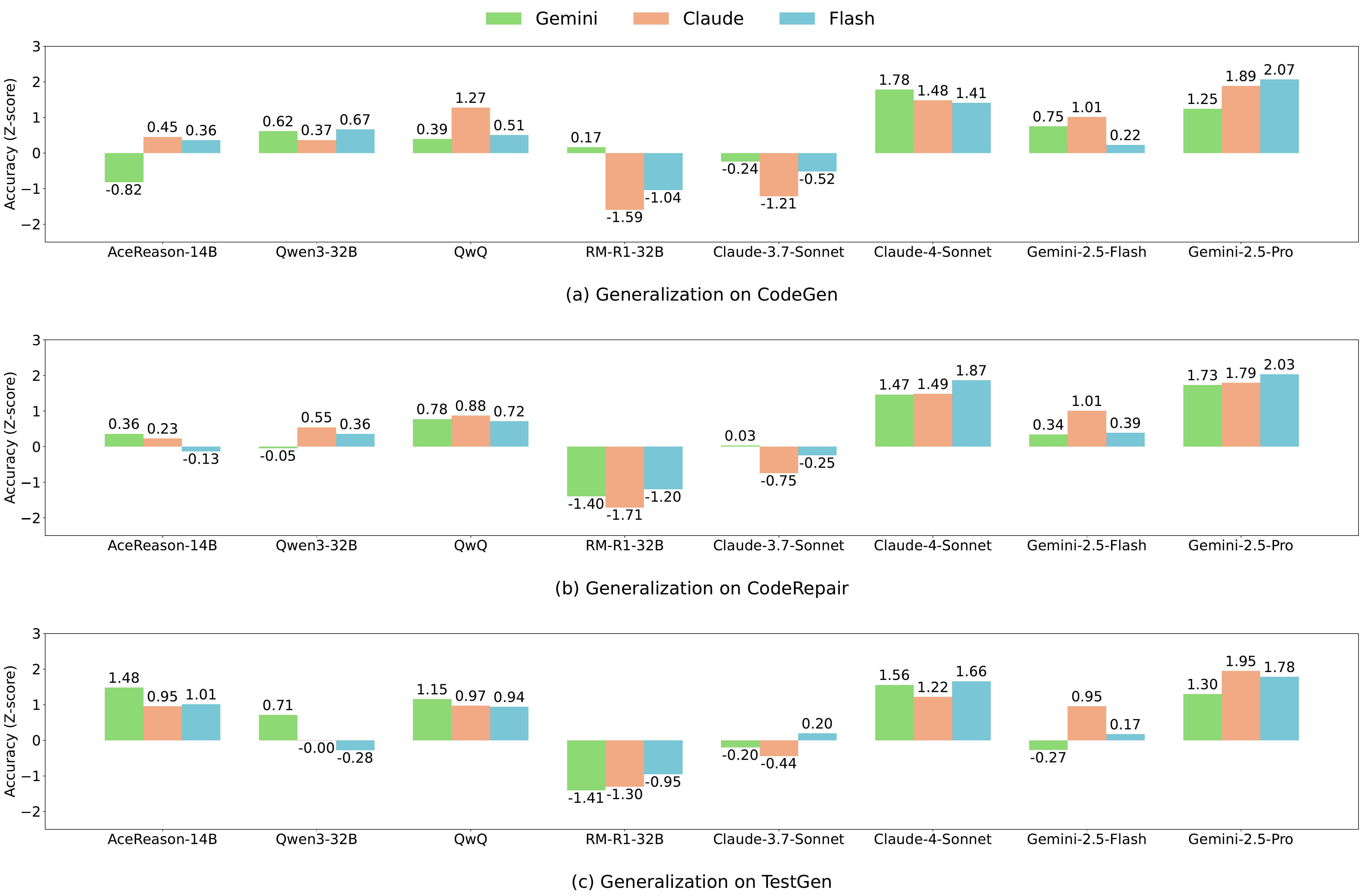}
    \caption{The performance of LLM-as-a-Judge on responses generated by Gemini-2.5-Pro (Gemini), Gemini-2.5-Flash (Flash), and Claude-3.7-Sonnet (Claude).}
    \label{fig:generalize}
\end{figure*}

\paragraph{\textbf{Response Ordering:}}
We first investigate whether LLM-as-a-Judge models produce consistent evaluations under trivial changes in response order, specifically by swapping the position of the correct response within the pair. Surprisingly, as shown in Fig.~\ref{fig:gen-study}, model performance varies substantially depending on the order, with discrepancies reaching up to 14\%. For certain models, this positional bias persists across all tasks: for example, RM-R1 32B and Claude 3.7 consistently exhibit recency bias, tending to prefer the response presented in the second position across CodeGen, CodeRepair, and TestGen tasks. In contrast, Qwen3-32B displays a task-dependent position bias, performing better when the correct response is first for CodeGen, but preferring the second position for CodeRepair. Gemini-2.5-Pro demonstrates the least position bias, suggesting that its judgments are based more on the substantive features of the responses rather than their order. In contrast, other models display greater variability and randomness, indicating a higher susceptibility to position effects.

\paragraph{\textbf{Different Coding Models:}}
In this experiment, we evaluate LLM-as-a-Judge models on responses generated by three different coding models: Gemini-2.5-Pro, Gemini-2.5-Flash, and Claude-3.7-Sonnet. 
To enable comparison across splits, we apply Z-score normalization to the accuracy of LLM-as-a-Judge models within each split. 
Ideally, LLM-as-a-Judge performance should remain consistent across outputs from different models, as code correctness is an objective criterion.
However, as shown in Fig.~\ref{fig:gen-study}, we observe significant variability in LLM-as-a-Judge performance on the different splits. 
For example, in the CodeGen task, QwQ is much better at judging responses from Claude-3.7-Sonnet than Gemini-2.5-Pro/Flash, while RM-R1-32B performs better on Gemini-2.5-Pro outputs. 
Unlike the response position bias, even Gemini-2.5-Pro does not exhibit consistent performance across different splits. These findings suggest that LLM-as-a-Judge models may not base their assessments solely on code correctness, but may also be influenced by additional factors such as coding style or response formatting. We further investigate response formatting in RQ3.

\begin{tcolorbox}[colback=paleviolet, size=title, breakable]
\textbf{\textcolor{titlecolor}{Answer to RQ2:}}
Overall, existing LLM-as-a-Judge models exhibit limited generalization capabilities. In particular, many models are highly sensitive to the ordering of responses in the pair-wise evaluation: their accuracy drops significantly depending on whether the good response is presented first or second. Stronger LLM-as-a-Judge models, such as Gemini-2.5-Pro, demonstrate greater robustness to such position swaps. Nevertheless, all models display considerable variability in performance when judging responses generated by different LLM Programmers. These findings highlight the importance of future research focused on improving the generalization and robustness of LLM-as-a-Judge systems.
\end{tcolorbox}

\clearpage
\subsection{RQ3: How does prompting impact LLM-as-a-Judge performance?}
\label{sec:rq3}

In this research question, we investigate how different prompting strategies affect the performance of LLM-as-a-Judge models. Specifically, we conduct three studies: (1) a comparison between point-wise and pair-wise evaluation schemes, (2) an analysis of how various pre-processing approaches applied to candidate responses influence judging accuracy, and (3) an assessment of Best-of-N using pair-wise prompting.

% \begin{table}[!htbp]
\begin{wraptable}{r}{0.55\textwidth}
\centering
{
\begin{tabular}{lccc}
\toprule \toprule
\textbf{LLM-as-a-Judge} & \textbf{Correct} & \textbf{Wrong} & \textbf{Tie} \\
\midrule
DeepCoder-14B & 27.1 & 16.62 & 56.28 \\
Phi4-Reasoning-Plus-14B & 38.18 & 16.82 & 45.0 \\
Qwen3-8B & 38.67 & 9.69 & 51.63 \\
Qwen3-14B & 45.5 & 7.62 & 46.88 \\
Qwen3-32B & 39.47 & 11.37 & 49.16 \\
QwQ-32B & 46.09 & 7.32 & 46.59 \\
R1-Distill-Qwen-14B & 23.84 & 15.73 & 60.44 \\
R1-Distill-Qwen-32B & 21.07 & 31.45 & 47.48 \\
R1-0528-Distill-Qwen3-8B & 32.54 & 30.76 & 36.7 \\
AceReason-Nemotron-14B & 41.84 & 12.36 & 45.8 \\
\bottomrule \bottomrule
\end{tabular}
}
\caption{The performance of point-wise prompting on CodeGen Task.}
\label{tab:point-wise-study}
\end{wraptable}
% \end{table}

\paragraph{\textbf{Point-wise vs. Pair-wise:}}
In addition to pair-wise evaluation, point-wise evaluation is another commonly used schema. In the point-wise approach, the LLM-as-a-Judge evaluates each candidate response independently, assigning a score on a scale from 1 to 5. The response with the highest score from the candidate pair is then selected as the preferred answer.
Our experiments on the CodeGen task show that the point-wise approach significantly underperforms compared to the pair-wise approach. Further analysis reveals that this discrepancy is primarily due to the frequent occurrence of tied scores between candidates. As shown in Tab.~\ref{tab:point-wise-study}, approximately 50\% of point-wise judgments result in ties for various models. We attribute this to the absence of direct comparison in the point-wise setting, making it difficult for the model to distinguish between highly similar candidates and resulting in arbitrary or indistinguishable scoring.
Moreover, since code evaluation is fundamentally a binary classification task, determining whether a solution is correct or not—rather than a subjective, fine-grained assessment. As such, the point-wise scheme is less suitable for this context. Consequently, we adopt the pair-wise prompting method in our experiments, as it consistently yields superior performance.

% \clearpage
\paragraph{\textbf{Candidate Pre-processing:}}
% \begin{figure}[h] 
\begin{wrapfigure}{r}{0.5\textwidth}
\centering
\includegraphics[width=0.5\textwidth]{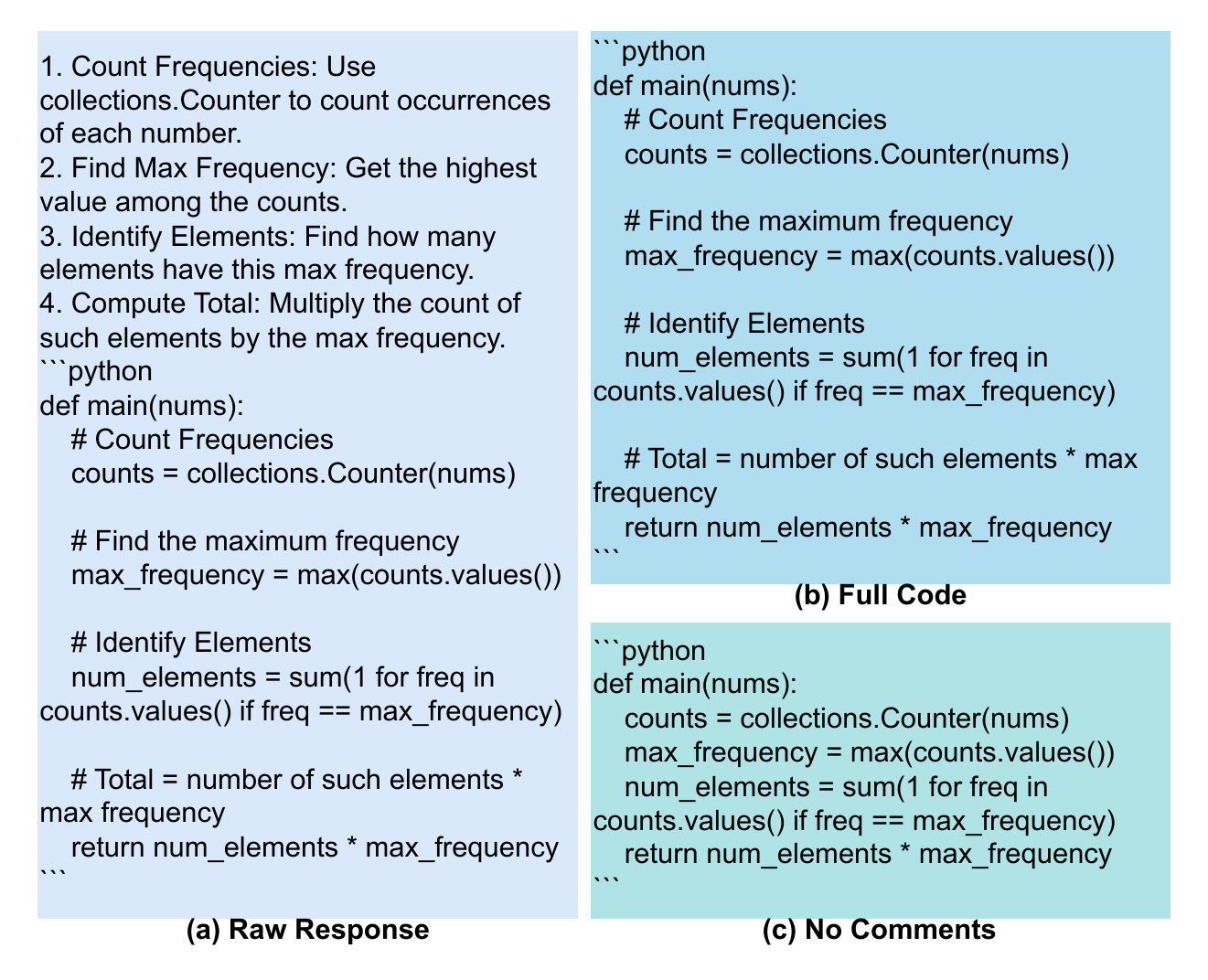}  
\caption{Illustration of response after different pre-processing.}
\label{fig:response-style}
% \end{figure}
\end{wrapfigure}
For the code generation task, the primary determinant of response quality is the generated code itself. 
Prior work~\cite{AceCoder} often applies post-processing to raw model outputs before passing them to LLM-as-a-Judge models. 
In this study, we systematically examine three variants of post-processing, as illustrated in Fig.~\ref{fig:response-style}. 
First, we consider the baseline approach with no pre-processing, using the raw model response as input to the judge. 
Second, we extract only the code segments contained within markdown code blocks, omitting any surrounding text. 
Third, we further refine the extracted code by removing all comments, retaining only the executable code.

The results are summarized in Tab.~\ref{tab:result-style}. We observe that removing comments from the code leads to a significant decline in LLM-as-a-Judge performance. Notably, in contrast to previous work that uses only code as input to the LLM-as-a-Judge, our findings indicate that providing the full model response, rather than code alone, consistently yields better performance on average.

\clearpage
\input{tables/style}
\paragraph{\textbf{CodeGen Best-of-N (BoN):}}
% \begin{table}[!htbp]
\begin{wraptable}{r}{0.51\textwidth}
\centering
\resizebox{0.5\textwidth}{!}{
\begin{tabular}{lcccc}
\toprule \toprule
\textbf{Source} & \multicolumn{3}{c}{\textbf{Num. Correct}} & \textbf{Overall} \\
\cmidrule(lr){2-4}
 & \textbf{4} & \textbf{3-2} & \textbf{1} &  \\
\midrule
Claude-3.7-Sonnet & 163 & 71 & 91 & 325 \\
Gemini-2.5-Flash & 267 & 90 & 73 & 430 \\
Gemini-2.5-Pro & 158 & 51 & 47 & 256 \\
\bottomrule \bottomrule
\end{tabular}
}
\caption{Data statistics of CodeGen BoN. ``Num. Correct" indicates the number of correct responses obtained when generating 5 candidate solutions per problem.}
\label{tab:bon_stats}
\end{wraptable}
% \end{table}
We investigate pair-wise prompting for BoN inference-time scaling on the CodeGen task, where the LLM-as-a-Judge must distinguish correct responses from a set of correct and incorrect candidates.
For each coding question, we sample 5 candidate responses and verify their correctness using unit tests.
Each correct response is then paired with all incorrect responses to create evaluation instances, and we alternate the position of the correct response within each pair to mitigate position bias. A summary of the BoN dataset is presented in Tab.~\ref{tab:bon_stats}.
Following the RMB~\cite{zhou2025rmb} evaluation protocol, an instance is correct only if the LLM-as-a-Judge always selects the correct response over all incorrect responses.

In Tab.~\ref{tab:bon}, we can see that closed-source models such as Gemini-2.5-Pro and Claude-4-Sonnet performs best. Overall, the model rankings is consistent with Tab.~\ref{tab:result:main}. Notably, AceReason-Nemotron-14B performs considerably worse than Qwen3-14B in the BoN setting, despite their similar performance in Tab.~\ref{tab:result:main}. This difference is likely due to its position bias as seen in Fig.~\ref{fig:gen-study}, which undermines its effectiveness when multiple comparisons are required.

\begin{tcolorbox}[colback=paleviolet, size=title, breakable]
\textbf{\textcolor{titlecolor}{Answer to RQ3:}}
We find that pair-wise evaluation is more suitable for coding-related tasks, which require more fine-grained analysis of candidate responses. However, inherent judgment biases in pair-wise comparisons underscore the importance of positional robustness when applying Best-of-N strategies. Additionally, our results indicate that providing LLM-as-a-Judge models with the entire raw response, without any pre-processing, leads to better performance. 
\end{tcolorbox}

%% file: tables/main_results.tex
\begin{table*}[!b]
\resizebox{\textwidth}{!}{
\begin{tabular}{lcccccccccccccccc}
\toprule \toprule
\textbf{LLM-as-a-Judge}  & \multicolumn{4}{c}{\textbf{Code Generation}}                       & \multicolumn{4}{c}{\textbf{Code Repair}}                  & \multicolumn{4}{c}{\textbf{Test Generation}}    & \textbf{Avg.}              \\
\cmidrule(lr){2-5} \cmidrule(lr){6-9} \cmidrule(lr){10-13}
                 & \textbf{Easy} & \textbf{Medium} & \textbf{Hard} & \textbf{Overall} & \textbf{Easy} & \textbf{Medium} & \textbf{Hard} & \textbf{Overall} & \textbf{Easy} & \textbf{Medium} & \textbf{Hard} & \textbf{Overall} & \\
\midrule
 Claude-3.5-Sonnet-v2     &  72.62 &    61.81 & 43.91  &       58.32 &  81.50 &    71.19 &  50.15 &       65.90 &  66.85 &    60.49 &  45.65 &       53.15 &     59.12 \\
 Gemini-2.0-Flash         &  62.25 &    56.47 & 43.37  &       53.21 &  70.19 &    59.96 &  46.95 &       57.87 &  65.22 &    52.16 &  49.29 &       53.33 &     54.80 \\
 Gemini-2.0-Flash-Lite    & 64.63  &    54.14 & 43.00  &       53.21 &  65.11 &    56.33 &  46.38 &       55.02 &  61.14 &    54.32 &  46.86 &       51.43 &     53.22 \\
 \midrule
 Claude-3.7-Sonnet        &  89.91 &    71.21 &  43.18 &       66.33 &  94.31 &    77.95 &  48.04 &       70.94 &  94.02 &    77.47 &  51.21 &       65.65 &     67.64 \\
 Claude-4-Sonnet          &  \textbf{98.63} &    88.53 &  56.27 &       79.15 &  99.12 &    \textbf{93.39} &  64.00 &       83.33 &  97.01 &    \textbf{92.90} &  64.88 &       77.32 &     79.93 \\
 Claude-4-Opus            & 97.12  &    84.91 &  51.81 &       75.89 &  \textbf{99.66} &    91.68 &  60.02 &       81.40 &  97.01 &    84.88 &  58.10 &       71.79 &     76.36 \\
 Gemini-2.5-Flash         & 96.90  &    85.09 &  55.07 &       77.15 &  98.17 &    87.55 &  53.20 &       77.00 &  91.03 &    80.56 &  58.40 &       69.82 &     74.66 \\
 Gemini-2.5-Pro           & 98.85  &    \textbf{90.00} &  \textbf{62.24} &       \textbf{81.98} &  99.32 &    \textbf{93.39} &  \textbf{67.82} &       \textbf{84.93} &  \textbf{97.83} &    89.81 &  \textbf{69.23} &       \textbf{79.46} &     \textbf{82.12} \\
 \midrule
 AceCodeRM-7B             & 59.51  &    52.07 &  \underline{50.66} &       53.97 &  37.40 &    46.37 &  51.34 &       45.62 &  70.11 &    55.56 &  56.28 &       59.17 &     52.92 \\
 AceCodeRM-32B            & 70.32  &    55.17 &  49.22 &       57.82 &  50.41 &    54.48 &  50.21 &       51.52 &  69.57 &    54.94 &  52.23 &       56.55 &     55.30 \\
 Qwen2.5-Coder-32B        & 76.37  &    62.33 &  41.80 &       58.87 &  77.57 &    64.79 &  46.23 &       61.25 &  61.14 &    54.01 &  47.06 &       51.49 &     57.20 \\
 Skywork-Critic-70B       & 72.77  &    63.28 &  45.66 &       59.46 &  71.21 &    61.52 &  48.61 &       59.30 &  61.14 &    45.37 &  45.95 &       49.17 &     55.98 \\
 Prometheus-14B           & 75.65  &    61.64 & 43.43  &       59.08 &  78.79 &    66.43 &  47.88 &       62.76 &  59.51 &    50.62 &  45.04 &       49.29 &     57.04 \\
 Self-Taught-70B          & 72.12  &    60.60 & 42.88  &       57.42 &  76.36 &    62.38 &  45.61 &       59.92 &  65.76 &    53.70 &  45.85 &       51.73 &     56.36 \\
 \midrule
 R1-0528-Distill-Qwen3-8B &  94.52 &    77.84 &  48.07 &       71.61 &  94.44 &    77.10 &  50.62 &       71.77 &  86.68 &    75.31 &  50.61 &       63.27 &     68.88 \\
 R1-Distill-Qwen-14B      & 86.96  &    71.03 &  40.23 &       64.15 &  92.68 &    78.17 &  44.37 &       69.03 &  89.40 &    70.06 &  47.37 &       60.95 &     64.71 \\
 R1-Distill-Qwen-32B      & 95.24  &    76.90 &  36.55 &       67.05 &  97.29 &    80.94 &  40.34 &       69.63 &  95.11 &    80.86 &  51.01 &       66.43 &     67.70 \\
 DeepCoder-14B            & 92.44  &    72.24 &  39.32 &       65.93 &  95.80 &    76.96 &  41.48 &       68.47 &  93.75 &    76.85 &  47.87 &       63.51 &     65.97 \\
 Qwen3-8B                 & 93.52  &    78.10 &  47.77 &       71.23 &  94.44 &    81.37 &  46.90 &       71.52 &  79.62 &    63.27 &  41.50 &       54.05 &     65.60 \\
 Qwen3-14B                & 98.05  &    83.02 &  43.49 &       72.40 &  98.78 &    88.55 &  43.65 &       73.64 &  95.92 &    79.94 &  52.83 &       67.50 &     71.18 \\
 Qwen3-32B                & 97.69  &    86.98 &  47.35 &       74.89 &  98.92 &    88.90 &  46.80 &       75.05 &  95.92 &    78.40 &  53.04 &       67.32 &     72.42 \\
 QwQ-32B                  & \underline{99.06}  &    \underline{89.74} &  49.34 &       \underline{76.89} &  \underline{99.53} &    \underline{90.68} &  \underline{52.89} &       \underline{78.21} &  97.83 &    \underline{92.28} &  59.31 &       74.11 &     \underline{76.40} \\
 Phi4-Reasoning-Plus-14B  & 92.44  &    74.57 &  46.56 &       69.42 &  82.18 &    66.86 &  44.68 &       62.64 &  89.40 &    79.01 &  54.15 &       66.67 &     66.24 \\
 AceReason-Nemotron-14B   & 97.12  &    82.50 &  45.90 &       72.90 &  98.10 &    86.06 &  47.26 &       74.16 &  \underline{98.91} &    89.20 &  \underline{61.03} &       \underline{74.76} &     73.94 \\
 RM-R1-14B                & 90.49  &    73.36 &  39.02 &       65.48 &  91.80 &    74.04 &  44.94 &       67.79 &  88.59 &    68.83 &  39.57 &       55.95 &     63.07 \\
 RM-R1-32B                & 92.07  &    74.05 &  37.94 &       65.76 &  89.84 &    71.19 &  38.79 &       63.89 &  84.24 &    73.77 &  43.32 &       58.15 &     62.60 \\
\bottomrule \bottomrule
\end{tabular}
}
\label{tab:result:main}
\caption{The performance of different judges on proposed CodeJudgeBench. Accuracy scores are reported separately for the easy, medium, and hard splits. Additionally, we report the average accuracy of three tasks. The first and second block show the non-thinking and thinking proprietary judges. The third and the fourth block show the non-thinking and thinking open-source judges. We highlight the best performance with \textbf{bold}, and the best open-source performance with \underline{underline}.}
\end{table*}

%% file: tables/style.tex
\begin{table*}[t]
\centering
\resizebox{\textwidth}{!}{
\begin{tabular}{lccccccccccc}
\toprule \toprule
\textbf{Model} & \multicolumn{3}{c}{\textbf{Gemini Split}} & \multicolumn{2}{c}{\textbf{Claude Split}} & \multicolumn{3}{c}{\textbf{Flash Split}} & \multicolumn{3}{c}{\textbf{Overall}} \\
\cmidrule(lr){2-4} \cmidrule(lr){5-6} \cmidrule(lr){7-9} \cmidrule(lr){10-12} 
 & \textbf{FC} & \textbf{NC} & \textbf{RR} & \textbf{FC} & \textbf{NC} & \textbf{FC} & \textbf{NC} & \textbf{RR} & \textbf{FC} & \textbf{NC} & \textbf{RR} \\
 \midrule
DeepCoder-14B & 56.25 & 55.47 & 55.86 & 75.85 & 74.77 & 61.74 & 60.93 & 64.19 & 64.61 & 63.72 & 65.30 \\
Qwen3-8B & 60.16 & 58.79 & 63.48 & 81.08 & 81.69 & 69.53 & 66.05 & 68.49 & 70.26 & 68.84 & 71.01 \\
Qwen3-14B & 62.50 & 61.52 & 66.02 & 81.38 & 82.46 & 73.26 & 68.02 & 74.53 & 72.38 & 70.67 & 73.98 \\
Qwen3-32B & 63.48 & 62.89 & 67.19 & 83.38 & 84.77 & 71.86 & 69.42 & 73.14 & 72.91 & 72.36 & 74.57 \\
QwQ & 61.33 & 63.28 & 66.21 & 88.15 & 87.08 & 72.33 & 72.91 & 72.21 & 73.94 & 74.42 & 75.52 \\
RM-R1-14B & 59.38 & 55.86 & 63.09 & 76.00 & 71.54 & 61.74 & 60.35 & 60.35 & 65.71 & 62.58 & 66.48 \\
RM-R1-32B & 58.98 & 55.86 & 65.23 & 73.08 & 72.00 & 62.91 & 59.77 & 63.26 & 64.99 & 62.54 & 67.19 \\
Skywork-Critic-70B & 57.03 & 56.05 & 64.06 & 57.69 & 58.15 & 62.56 & 60.35 & 61.16 & 59.09 & 58.19 & 60.97 \\
Gemini-2.5-Pro & 72.85 & 71.88 & 69.92 & 90.46 & 90.31 & 82.79 & 81.98 & 81.28 & 82.03 & 81.39 & 80.55 \\
Gemini-2.5-Flash & 63.09 & 59.77 & 67.77 & 86.00 & 84.77 & 69.65 & 69.88 & 70.58 & 72.91 & 71.47 & 74.78 \\
Claude-3.7-Sonnet & 61.33 & 58.20 & 63.48 & 73.38 & 74.15 & 66.16 & 65.00 & 66.28 & 66.96 & 65.79 & 67.71 \\
Claude-4.0-Sonnet & 68.95 & 69.92 & 72.27 & 87.38 & 89.23 & 77.56 & 78.72 & 77.44 & 77.96 & 79.29 & 79.03 \\
\midrule
Overall & 62.11 & 60.79 & 65.38 & 79.49 & 79.24 & 69.34 & 67.78 & 69.41 & 70.31 & 69.27 & 71.43 \\
\bottomrule \bottomrule
\end{tabular}
}
\caption{The performance of different LLM-as-a-Judge models on the proposed CodeJudgeBench under various input pre-processing strategies. Note that Claude typically generates code-only responses, making the raw response identical to the full code output. Therefore, for the Claude split, we report only the full code and no comments results. FC refers to full code, NC refers to no comments, and RR refers to raw response.}
\label{tab:result-style}
\end{table*}

%% file: texs/relatedwork.tex
\section{Related Work}
\input{tables/bon}

Instruction fine-tuned~\cite{ouyang2022training} models demonstrate the ability to perform a wide range of tasks~\cite{huang2023empirical,deng2023pentestgpt,wu2023bloomberggpt,yang2023fingpt,li2024llms}, owing to emergent capabilities at scale~\cite{wei2022emergent}. As a result, these models can be directly prompted to judge responses without additional task-specific training. For example, Zheng et al.~\cite{zheng2023judging} found that using GPT-4 as a judge yields high correlation with human evaluations.
There are two primary prompting strategies for eliciting judgments from LLMs: pairwise grading, where the model compares two responses, and single-answer grading, where each response is evaluated independently. Including a reference response can further anchor the model's evaluation, making the judgment process more objective. In the absence of such a reference, the LLM must rely solely on its internal knowledge, which can introduce inconsistencies. Prior work~\cite{zheng2023judging,tan2025judgebench} has shown that better results are achieved when the LLM is first asked to generate a correct answer to serve as a reference.
Prometheus~\cite{kim2024prometheus} further proposed augmenting prompts with detailed rubrics, while other approaches, such as PandaLM~\cite{wang2024pandalm}, leverage training on human rationales and preference data to enhance evaluation quality.

LLM Judge shows certain biases, such as position bias, style bias, and length bias. They tend to prefer verbose answers and well-formatted responses.
\cite{zheng2023judging} found that when judging incorrect responses, the LLM tends to make similar mistakes which means it could be misled by the response. 
To improve the LLM's judging ability, recent work uses supervised fine-tuning to either try to mitigate certain biases or improve the ability to use the reference response. JudgeLM~\cite{zhu2025judgelm} fine-tuned an LLM judge to mitigate certain response biases. To better utilize the LLM's generative capabilities, some LLM Judges use chain-of-thoughts (CoT) to break down their judgment process and to give an explanation of the final judgment.
Auto-J~\cite{li2024generative} is fine-tuned on critiques generated by GPT4.
CritiqueLLM~\cite{ke2023critiquellm} used an advanced prompting technique to elicit more fine-grained judgment from GPT4 by asking it to critique each response individually before combining the individual critiques into a fine-grained pair-wise comparison.

In coding tasks, LLM judges are typically off-the-shelf models, though recent efforts have aimed to develop specialized coding judges. For example, AceCodeRM~\cite{AceCoder} is a point-wise discriminative LLM judge trained on the specially curated AceCoder-89K dataset, demonstrating strong potential in reinforcement learning and test-time scaling scenarios. Similarly, CriticGPT~\cite{mcaleese2024llm} is an LLM designed to detect bugs in code, trained via RLHF with a reward model that can assess bug severity. Despite these advancements, there remains a clear need for stronger and more comprehensive benchmarks to drive progress in LLM-based code judging.
General LLM-as-a-Judge benchmarks~\cite{tan2025judgebench,liu2025rmbench} include a coding split, but these are typically small in scale and focus primarily on code generation. In coding-specific evaluations, recent works~\cite{yang2025code0diting0,ficek2025scoring,zhou2025evaluating} have mainly focused on judging basic code generation tasks sourced from MBPP~\cite{austin2021program} and HumanEval~\cite{chen2021evaluating}, which feature limited algorithmic complexity. To address these limitations, we introduce CodeJudgeBench, which substantially advances the evaluation of LLM judges in coding by providing challenging data points across three critical tasks: code generation, code repair, and unit test generation.

%% file: tables/bon.tex
\begin{table}[!htbp]
% \begin{wraptable}{r}{0.5\textwidth}
\centering
\resizebox{0.5\textwidth}{!}{
\begin{tabular}{lcc}
\toprule \toprule
\textbf{LLM-as-a-Judge} & \multicolumn{2}{c}{\textbf{Coding Model}} \\
\cmidrule(lr){2-3}
 & \textbf{Gemini-2.5-Pro} & \textbf{Claude-3.7-Sonnet} \\
\midrule
Claude-3.7-Sonnet & 29.85 & 46.00 \\
Claude-4-Sonnet & 44.31 & 73.46 \\
Claude-4-Opus & 40.92 & 64.99 \\
Gemini-2.5-Flash & 37.54 & 73.68 \\
Gemini-2.5-Pro & 42.46 & 81.92 \\
\midrule
R1-0528-Distill-Qwen3-8B & 30.77 & 60.18 \\
R1-Distill-Qwen-14B & 26.15 & 52.17 \\
R1-Distill-Qwen-32B & 26.46 & 57.67 \\
DeepCoder-14B & 21.54 & 49.43 \\
Qwen3-8B & 30.77 & 60.87 \\
Qwen3-14B & 37.23 & 65.22 \\
Qwen3-32B & 34.77 & 66.82 \\
QwQ-32B & 37.54 & 73.68 \\
Phi4-Reasoning-Plus-14B & 33.23 & 60.87 \\
AceReason-Nemotron-14B & 29.85 & 61.10 \\
RM-R1-14B & 32.31 & 52.63 \\
RM-R1-32B & 39.38 & 51.03 \\

\bottomrule \bottomrule
\end{tabular}
}
\caption{Comparison of LLM-as-a-Judge performance using pair-wise prompting for CodeGen BoN.}
\label{tab:bon}
% \end{wraptable}
\end{table}

%% file: texs/conclusion.tex
\section{Conclusion}

In this work, we introduce CodeJudgeBench, a benchmark designed to evaluate LLM-as-a-Judge models across a variety of coding tasks. Through a comprehensive evaluation of 26 LLM-as-a-Judge models, we confirm the strong performance of recent thinking judges and highlight this as a promising research direction. Nevertheless, our extensive analysis reveals that the robustness and generalization capabilities of current LLM-as-a-Judge models still require significant improvement. In addition, we find that pair-wise evaluation using full model responses constitutes a more effective design choice for LLM-as-a-Judge systems. Future work will focus on expanding CodeJudgeBench by incorporating additional tasks and continually updating the evaluation dataset with newly released coding problems.